%% file: baspinar2021Completion_arXivVersion.tex
\documentclass[article,onefignum,onetabnum]{siamart190516}

\input{ABD_shared}

\ifpdf
\hypersetup{
  pdftitle={Multi-frequency image completion via a biologically-inspired sub-Riemannian model with frequency and phase},
  pdfauthor={}
}
\fi

\usepackage[ruled, vlined]{algorithm2e}

\usepackage{ulem}

\usepackage{color}

\usepackage{mathtools}
\usepackage{environ}
\usepackage{amsmath}
\usepackage{amsfonts}
\usepackage{amssymb}
\usepackage[utf8]{inputenc}
\usepackage{caption}
\usepackage{subcaption}

\usepackage{todonotes}

\usepackage{placeins}
\usepackage[margin=3cm]{geometry}
\usepackage[belowskip=-5pt,aboveskip=5pt]{caption}

\newcommand{\se}{\text{SE}(2)}

\graphicspath{{figures/}}

\begin{document}

\maketitle

\begin{abstract}
We present a novel cortically-inspired image completion algorithm. It uses a five dimensional sub-Riemannian cortical geometry modelling the orientation, spatial frequency and phase selective behavior of the cells in the visual cortex. The algorithm extracts the orientation, frequency and phase information existing in a given two dimensional corrupted input image via a Gabor transform and represent those values in terms of cortical cell output responses in the model geometry. Then it performs completion via a diffusion concentrated in a neighbourhood along the neural connnections within the model geometry. The diffusion models the activity propagation integrating orientation, frequency and phase features along the neural connections. Finally, the algorithm transforms back the diffused and completed output responses back to the two dimensional image plane.
\end{abstract}


\section{Introduction}

\graphicspath{{figures/}}

Visual perception has drawn attention of experts from philosophy, psychology, neuroscience, as well as the attention of mathematicians and physicist working on perceptual modeling. The question of how we perceive was studied by Edmund Husserl in his pioneering philosophical texts in phenomenology \cite{husserl1968logische, husserl1991ding, husserl2009ideen}. In the side of psychology, we can think of the well known Berlin school of experimental psychology, \textit{Gestalt psychology school}, which formulated what is known today as \textit{Gestalt psychology of perception} \cite{wertheimer1938laws, kohler1970gestalt, koffka2013principles}.

Gestalt psychology is a theory which attempts to provide the principles giving rise to a meaningful global perception of a visual scene by starting from the local properties of the objects within the scene. The main idea of Gestalt psychology is that the mind constructs the global whole by rather grouping similar fragments than purely summing the fragments as if they were indifferent. In terms of visual perception, those similar fragments can be thought of as the point stimuli with the same (or closely) valued features of the same type. In the present paper, we consider orientation, spatial frequency and phase as features. We extract these features from a given two dimensional grayscale input image which is partially occluded and reveal those occluded parts via an integration of the extracted features. By doing so, we follow the Gestalt principle which is called \textit{the law of good continuity} to provide a feature integration mechanism reconstructing the occluded parts in the image. 

\begin{figure}[htp]
\centerline{\includegraphics[scale=0.25,trim={0cm 0 0 0},clip]{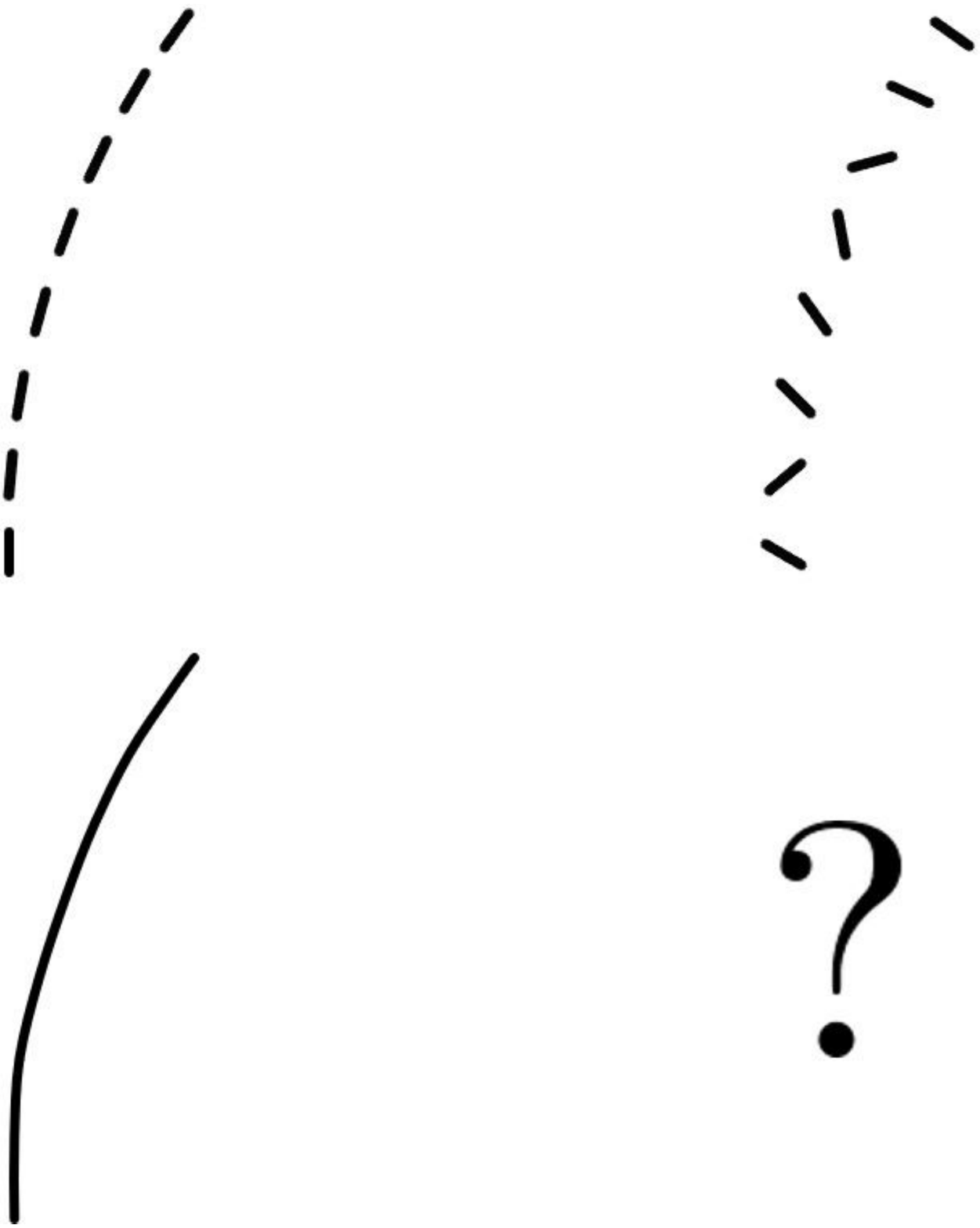}}
\caption{An example of the law of good continuity. We capture the curve on the bottom left as the curve underlying the aligned fragments on the top left, we do not capture any curve underlying the fragments on the top right due to the abruptly changing orientation angles of the fragments.}
\label{fig:goodCont}
\end{figure}

\begin{figure}[htp]
\centerline{\includegraphics[scale=0.2,trim={0cm 0 0 0},clip]{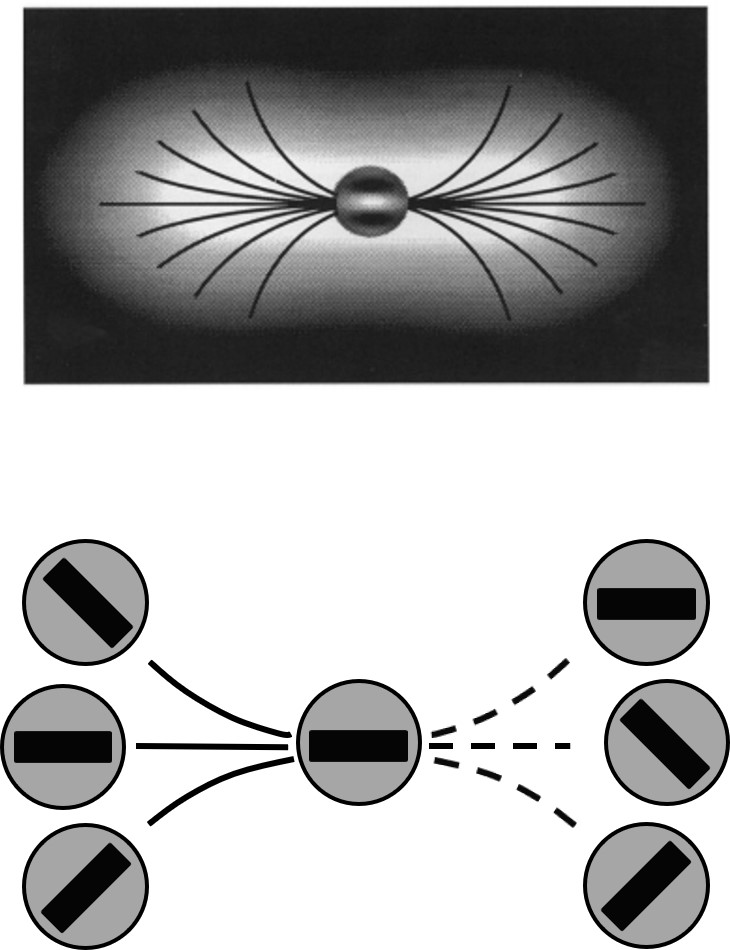}}
\caption{Top: association fields aligned with a horizontal patch as shown by Field, Hayes and Hess \protect\cite{field1993contour}. Bottom: solid curves represent the association fields between strongly associated fragments, and the dashed ones imply no fields between weakly associated fragments.\bigskip}
\label{fig:associationFields}
\end{figure}

\begin{figure}[htp]
\centerline{\includegraphics[scale=0.35,trim={0cm 0 0 0},clip]{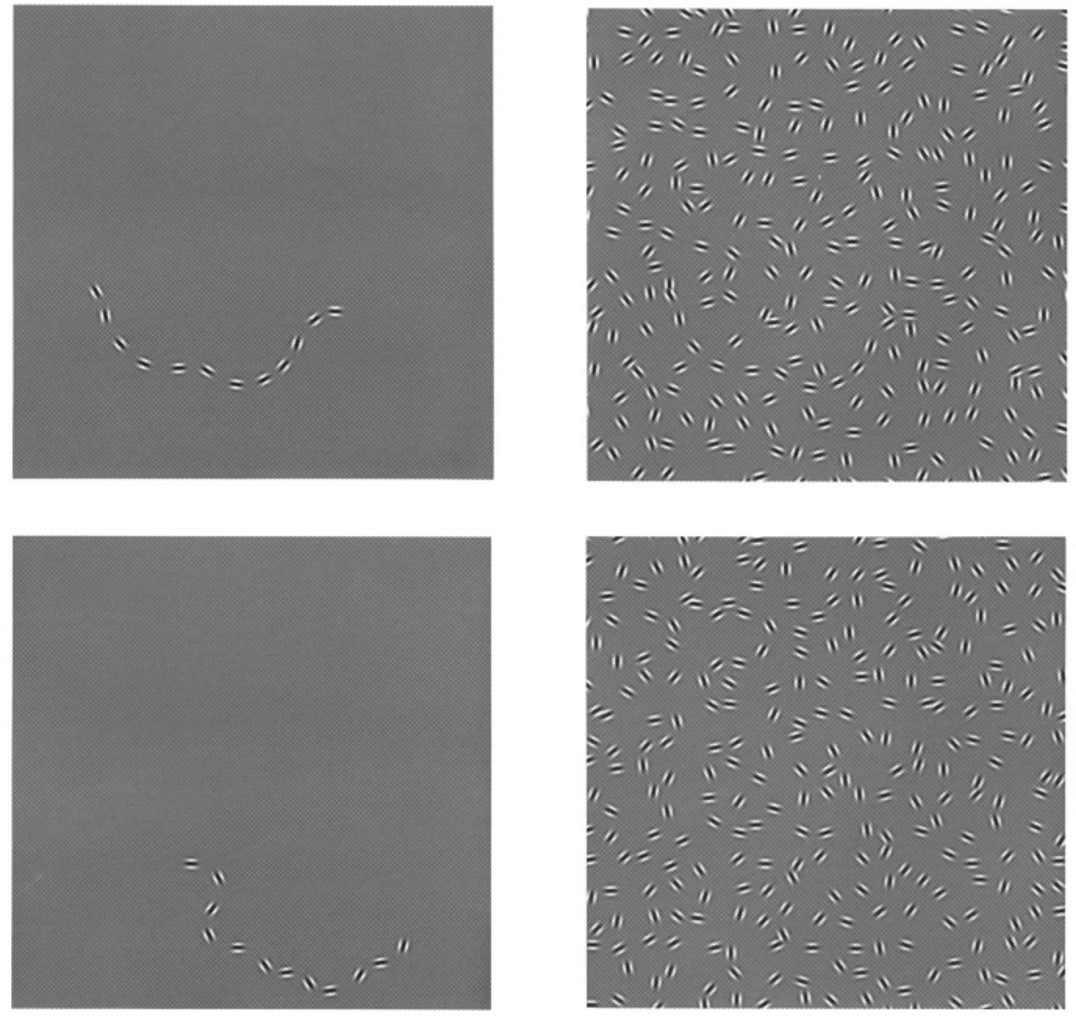}}
\caption{Two experimental settings from Field, Hayes and Hess \protect\cite{field1993contour}. A stimuli with aligned patches which we capture (left) and a stimuli plus the background with randomly oriented patches (right) are shown. Abrupt changes in the fragment orientations make it difficult to detect the aligned pattern in the bottom row.}
\label{fig:fieldExp}
\end{figure}

The law of good continuity states that we group rather aligned pieces than those with sharp abrupt directional changes when we perceive an object as a whole which is formed by fragments, see Figure~\ref{fig:goodCont}.
This law was studied by Field, Hayes and Hess \cite{field1993contour} from a psychophysical point of view. They worked on how the visual system captures aligned fragments constituting lines or cocircular curves on a two dimensional background of randomly oriented fragments; see Figure~\ref{fig:fieldExp}. They noticed that the aligned patterns captured by the visual system were locally overlapping with what they called \textit{association fields}, which are shown in Figure~\ref{fig:associationFields}. Those fields provide a geometric characterization of the law of good continuity and they can be interpreted as the psychophysical representations of the neural connections which are biologically implemented in the visual cortex.

The visual system is capable of perceiving a lacunar curve as complete by capturing the whole curve pattern underlying the lacunar curve; see Figure~\ref{fig:goodCont}. This is due to a general 
phenomena which is called \textit{perceptual completion}. This phenomena provides the perception of contours and figures which are  actually not present in the visual stimulus. We call such contours \textit{subjective contours}.  

\begin{figure}[htp]
\centerline{\includegraphics[scale=0.2,trim={0cm 0 0 0},clip]{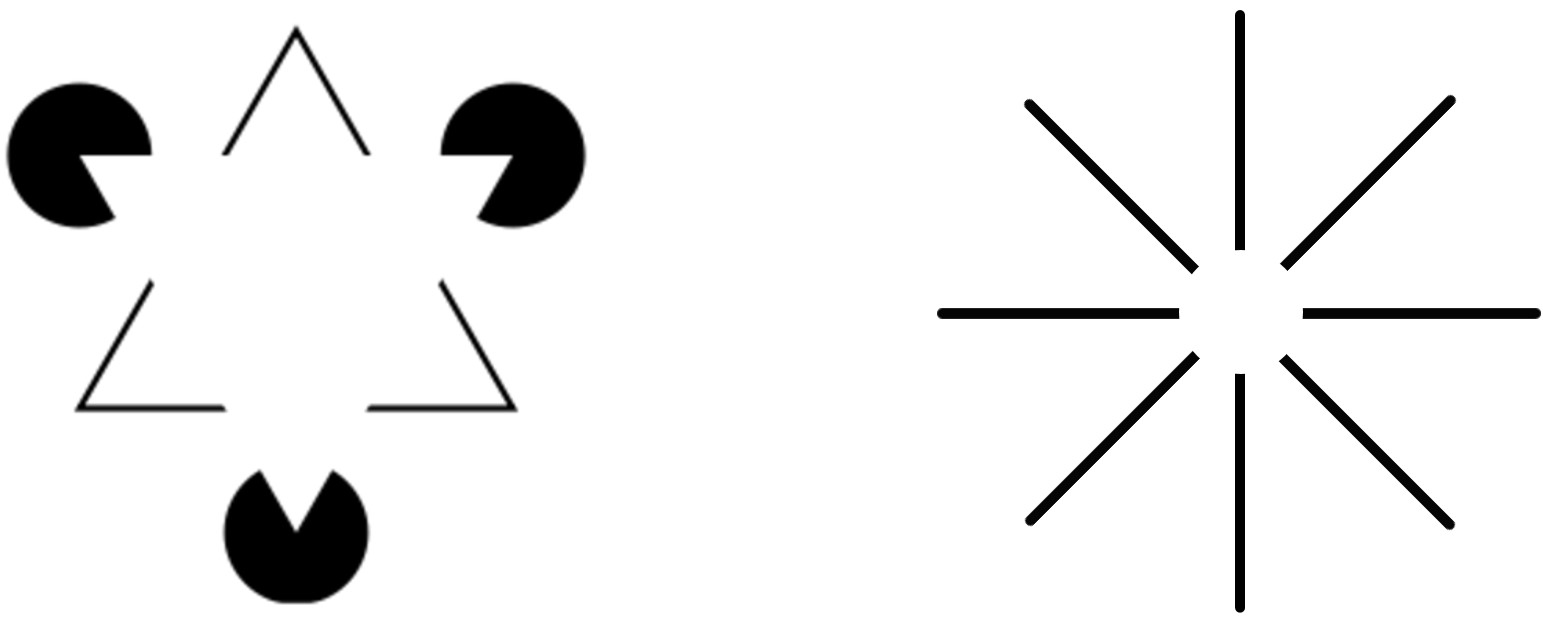}}
\caption{Left: Kanizsa triangle. There is no direct stimulus, yet we perceive a white triangle on top of the rest (modal completion). We perceive another triangular whose border is marked by the black lines on the bottom layer (amodal completion). Right: Ehrenstein illusion. We perceive a white disk around the center despite the absence of a direct stimulus (modal completion). We recognize that each vertical, horizontal or diagonal line fragment comprises a whole line which is occluded by the white disk (amodal completion).}
\label{fig:KanizsaEhrenstein_illusory}
\end{figure}

Kanizsa \cite{kanizsa1979organization, kanizsa1980grammatica} explains two categories of completion: modal completion and amodal completion. The first one refers to the completion following the modality of vision. There is no direct stimulus corresponding to the object, yet, the perceived object is indistinguishable from the real stimuli and we perceive it as a whole. In the second category, the completion does not make use of the modality of vision. In other words, the stimulus corresponding to the object is partial but we still perceive the object as complete. In this case, the object is recognized as a whole although only some specific fragments of the object evoke our sensory receptors. Examples of those two categories of completion resulting in subjective contours are Kanizsa triangle and Ehrenstein's illusion, which are illustrated  in Figure~\ref{fig:KanizsaEhrenstein_illusory}. We focus rather on amodal completion in the present work, and provide an approach which reconstructs the occluded parts in a compatible way with the law of good continuity and association fields, where those two notions are considered in an extended way based on position, orientation, frequency and phase alignment. 

The primary visual cortex (V1) is the main area of the cerebral cortex which is responsible for the first step processing of visual input so that a proper visual perception is achieved at a higher perceptual level. V1 contains a particular family of neurons, \textit{simple cells}. These neurons are locally sensitive to visual features, such as orientation \cite{hubel1959receptive, hubel1962receptive, hubel1963shape, hubel1977ferrier}, spatial frequency \cite{maffei1977spatial, hubener1997spatial, issa2000spatial, issa2008models, sirovich2004organization, tani2012parallel,  ribot2013organization, ribot2016pinwheel},  phase \cite{de1983spatial, pollen1988responses, levitt1990spatio, mechler2002detection}, scale \cite{blakemore1969existence} and ocular dominance \cite{shatz1978ocular, levay1978ocular, issa2000spatial}. The simple cells are organized in a {hypercolumnar architecture}, which was first discovered by Hubel and Wiesel \cite{hubel1974uniformity}. In this architecture, a hypercolumn is assigned to each point $(x,y)$ of the retinal plane $M\simeq \R^2$, and the hypercolumn contains all the simple cells sensitive to a particular value of the same feature type. The simple cells are able to locally exctract the feature values of the visual stimulus, and the activity propagation along the neural connections between the simple cells integrates those values to a coherent global unity. Those two mechanisms, the feature detection and the neural connectivity, comprise the functional geometry of V1.

A simple cell is identified by its receptive field, which is defined
as the domain of the retina to which the cell is sensitive and
connected through the retino-geniculo-cortical paths. Once
a receptive field is stimulated, it evokes a spike transmitted to the corresponding simple cells. Each one of those simple cells produces a response to the spike. This response is what is known as \textit{receptive profile}. 

Hoffman proposed modelling the hypercolumnar architecture of V1 in terms of a contact bundle structure \cite{hoffman1970higher, hoffman1989visual}. This framework was followed by Petitot and Tondut, where they improved the contact bundle structure and proposed a boundary completion method within the corresponding contact geometry \cite{petitot1999vers}. Moreover, they geometrically interpreted the association fields as the integration along the vector fields generating the contact geometry. This setting was developed further by Citti and Sarti \cite{citti2006cortical} to a framework in which they introduced a group based approach to study the geometric modeling of V1 hypercolumnar architecture and the functional connectivity. They used \text{Gabor function} as the receptive profile model and proposed the sub-Riemannian geometry of the group of rotations and translations ($\se$) as the V1 model geometry. This framework was extended to higher dimensional geometries where scale \cite{sarti2008symplectic}, and velocity \cite{barbieri2014cortical, cocci2015cortical} were taken into account. Various biologically-inspired models for optical illusions~\cite{franceschiello2017modelling, franceschiello2018neuromathematical, bertalmio2020visual, bertalmio2021cortical, baspinar2021cortical} and orientation preference maps~\cite{baspinar2018geometric}, as well as frameworks for image processing~\cite{duits2010left1, duits2010left2, boscain2012anthropomorphic, duits2013evolution, boscain2014image, sharma2015left, citti2016sub}, for pattern recognition~\cite{PGbook} and for medical applications \cite{ter2016brain, bekkers2014multi}, were proposed in the sub-Riemannian geometry of $\se$.
 
\begin{figure}[htp]
\centerline{\includegraphics[scale=0.75,trim={0cm 0 0 0},clip]{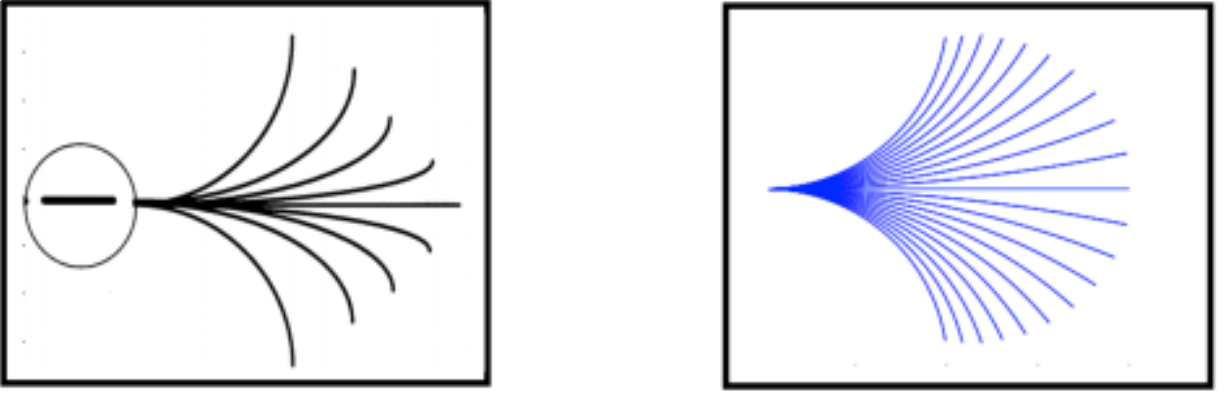}}
\caption{Left: real association fields. Right: projections of $\se$ horizontal integral curves. Figures are adapted from~\cite{field1993contour,citti2006cortical}.}
\label{fig:associationFieldsIntegral}
\end{figure}

The model presented in \cite{citti2006cortical} is an abstract geometric description of the orientation-sensitive V1 hypercolumnar architecture reported by Hubel and Wiesel \cite{hubel1959receptive, hubel1962receptive, hubel1963shape} . This description provides a good phenomenological approximation of the biologically implemented V1 neural connections which were reported by Bosking et al. \cite{bosking1997orientation}. In this model framework, the projections of a particular family of curves onto the two dimensional image plane provide good approximations of the association fields. In other words, these curves model the V1 neural connections; see Figure~\ref{fig:associationFieldsIntegral}. They are called \textit{horizontal integral curves} and they are obtained by integration along the vector fields generating the $\se$ sub-Riemannian geometry. For this reason, the approach considered by Citti, Petitot and Sarti and used in our present paper is referred to as \textit{biologically-inspired}.

The sub-Riemannian model geometry proposed in \cite{citti2006cortical} was extended in a recent work to multi-frequency and multi-phase setting \cite{baspinar2020sub}. The extended model corresponds to a natural geometry which is \textit{derived} \cite{baspinar2018minimal, baspinar2020sub} from one of the very first perceptual mechanisms of vision: receptive profile. It is different from the classical approach, in which a suitable geometry is \textit{assigned} to the neural responses represented in terms of receptive profiles. This extended model takes advantage of orientations, spatial frequencies and phases in a given 2D input image to encode the visual information. This is not the case in the sub-Riemannian model proposed in \cite{citti2006cortical}, which uses the $\se$ geometry, and in which only orientation can be represented as visual feature. The extended model geometry was applied to enhancement on images in which several spatial frequencies were equally present \cite{baspinar2020sub}, such as texture images. In the present work, we employ the same extended multi-frequency sub-Riemannian setting presented in \cite{baspinar2020sub}, and propose an image completion algorithm within, which is aimed for grayscale texture images containing multiple spatial frequencies.

In Section~\ref{sec:model}, we explain the model framework which was introduced in \cite{baspinar2020sub} and its relation to our completion algorithm. In Section~\ref{sec:Connectivity_in_the_extended_phase_space}, we present a specific family of integral curves defined in the model geometry, which are the models of the neural connections in V1. Then in Section~\ref{sec:algorithm}, we introduce our completion algorithm, its discrete scheme and the corresponding pseudocode. Finally, we provide our simulation results and their comparison to some results obtained previously in \cite{boscain2014hypoelliptic} and \cite{citti2016sub}. At the end, we give the main conclusions and some perspectives for the related future research.

\section{Model framework}\label{sec:model}
In this section, we will explain the cortical model geometry in which our completion algorithm is defined. This model framework is composed by two mechanisms: feature value extraction and neural connectivity \cite{baspinar2020sub}.

\subsection{Feature value extraction}
Each simple cell is sensitive to a specific part of the retina, which is called \textit{receptive field}. Once the receptive field is stimulated by visual stimulus, the retinal cells in the receptive field produce spikes which are transmitted through retino-geniculo-cortical pathways to the related simple cells in V1. Each simple cell generates a response to those spikes, which is the receptive profile corresponding to the simple cell. In other words, receptive profile is the impulse response function of the simple cell. The simple cell receptive profile which is sensitive to the stimulus located at $q\in M$ on the image plane $M$ and selective to the set of feature values $z\in S^1 \times\mathbb{R}^+ \times S^1$ is denoted by $\Psi_{(q,z)}$, where $q=(q_1,q_2)$ and $z=(\theta, f, \phi)$ together denote a fixed point $(q,z)$ in $Q\simeq \mathbb{R}^2\times S^1 \times\mathbb{R}^+ \times S^1$. Here $Q$ represents the five dimensional sub-Riemannian V1 model geometry. 

\begin{figure}[htp]
\centerline{\includegraphics[scale=0.7,trim={0cm 0 0 0},clip]{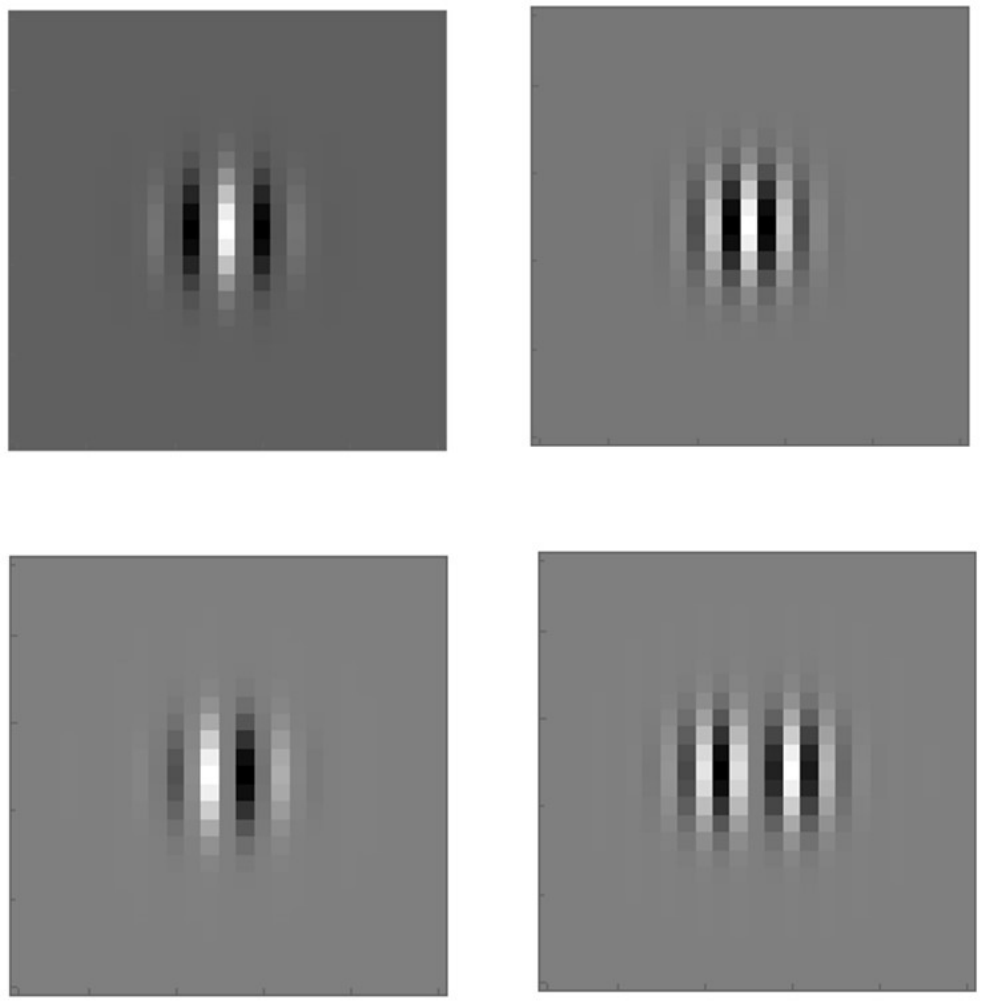}}
\caption{Two Gabor functions with low (left column) and high (right column) spatial frequencies. Top row: even (real) component of the Gabor functions. Bottom row: odd (imaginary) components.}
\label{fig:gabor}
\end{figure}

Simple cell receptive profiles can be modeled in terms of Gabor functions \cite{daugman1985uncertainty, citti2006cortical, baspinar2020sub, baspinar2021cortical}. In the orientation, frequency and phase selective model framework, receptive profile of a simple cell is a Gabor function of the following type:
\begin{equation}\label{eq:gaborExtendedFunctionExtended}
\Psi_{(q,z)}(x,y,s):=\frac{1}{2\,\sigma^2}\expp^{-i\big(r \cdot (x-q_1,\, y-q_2)-(s-\phi)\big)}\expp^{-\frac{\lvert x-q_1\rvert^2-\lvert y-q_2\rvert^2}{2\,\sigma^2}},
\end{equation}
where $f>0$ represents the spatial frequency\footnote{Spatial frequency is found via $f=\frac{1}{\lambda}$, where $\lambda>0$ denotes the wavelenght.}, $r=(r_1, r_2)=(-f\sin\theta,\,f\cos\theta)$ and $\sigma>0$ is the scale of the localizing Gaussian. The complex exponential is the wave content and it is the main component capturing the orientation, frequency and phase information of the objects in a given two dimensional image. The second exponential is a Gaussian which spatially localizes the receptive profile around the point $(q_1, q_2)$. Frequency $f$ determines how many wave peaks are found within the localizing Gaussian window; see Figure~\ref{fig:gabor}. As the number of wave peaks increases, the Gabor function can detect higher frequencies. Orientation $\theta$ is the orientation angle to which the simple cell receptive profile is sensitive. Parameter $\phi$ is the reference phase and it creates a phase shift in the waves of the Gabor function as it changes. 

We disregard the coordinate map between the image plane and the retinal surface, as well as the retino-cortical map between the retinal surface and the V1 surface. We assume that the image plane is identically mapped to the retinal and cortical surfaces. We assume the responses of the simple cells to be linear and we compute the output response of a simple cell located at $(q, z)\in Q$ to a given two-dimensional grayscale image $I:M\rightarrow [0,1]$ via the convolution with the Gabor filter:
\begin{align}\label{eq:liftingConv}
O_I(q,z) =  \int_M \Psi_{(q, z)}(x,y, 0)I(x,y)\,dx\,dy.
\end{align}    
We apply the convolution for every feature value $z$ and for every point $q$. Consequently, we obtain the output responses of all receptive profiles corresponding to the V1 simple cells. We will call this set of output responses sometimes, \textit{lifted image}; and the Gabor transform, \textit{lifting}. It is equivalent to the result of a multi-frequency Gabor transform applied to the given two dimensional input image. Those output responses are the representations of the feature values in the five dimensional V1 model geometry $Q$.

In general, static receptive profile models based on linear filter banks and static nonlinearities \cite{koenderink1987representation, citti2006cortical, duits2007invertible, lindeberg2013computational, bekkers2018roto, baspinar2020sub} provide good responses to simple stimuli. However, their responses to complicated stimuli, such as natural images, are approximate up to a certain level. Several mechanisms such as response normalisation, gain controls, cross-orientation suppression and intra-cortical modulation, can result in radical changes in the receptive profile shape. Therefore, the aforementioned Gabor filter bank model for the receptive profiles should be considered as a first approximation of highly complex real dynamic receptive profile.

We employ all frequency components of the Gabor transform during the lifting. Therefore, exact inverse Gabor transform is valid and we use it to obtain the corresponding two dimensional image to the output responses:
\begin{equation}\label{eq:inverseGaborTransformExpression}
I(q_1, q_2)=\frac{\sqrt{f}}{\| \Psi  \|_{\mathbb{L}^2}}\int_{Q} O_I(x,y,z)\bar{\Psi}_{(x,y,z)}(q_1, q_2, 0)\,dx\,dy\,dz,
\end{equation}
with $\bar{\Psi}$ denoting the complex conjugate of the corresponding Gabor function $\Psi$. 

\subsection{Horizontal connectivity}
Lifting provides the output responses, which are complex valued functions in the five dimensional model geometry $Q$. Each output response encodes the feature values corresponding to the orientation, frequency and phase of a pixel defined on the two dimensional image plane. The output responses, hence the simple cells, are isolated from each other once lifting from the image plane $M$ to the model geometry $Q$ takes place. Therefore, the model geometry $Q$ should be endowed with an integration mechanism which provides the activity propagation, therefore the interactiviy, between the simple cells. The activity propagation provides an integrated form of the local feature vectors associated to the lifted image. This propagation is concentrated along a specific family of integral curves, \textit{horizontal integral curves}, corresponding to the model geometry $Q$. The horizontal integral curves can be thought of as the models of the long range lateral connections, which connect the simple cells residing in different hypercolumns but selective to the same (or close) feature values.

We may associate the following differential one-form to each receptive profile described by \eqref{eq:gaborExtendedFunctionExtended}:
\begin{equation}
\Theta_{(\theta, f)} = -f\sin(\theta)\,dx + f \cos(\theta)\,dy - ds.
\end{equation}
The one-form induces naturally the \textit{horizontal vector fields} corresponding to the model geometry $Q$. The horizontal vector fields are formally defined as the elements of
\begin{equation}\label{eq:oneForm}
\operatorname{ker}(\Theta) : = \{X\in TQ: \;\Theta(X) = 0     \},
\end{equation}
where $TQ$ denotes the tangent bundle of $Q$. The horizontal vector fields corresponding to $Q$ are found from \eqref{eq:oneForm}  as
\begin{equation}\label{eq:LieAlgebra}
 \begin{split}
 X_1 = & \cos(\theta)\,\partial_x + \sin(\theta)\, \partial_y,\quad\quad X_2 = \partial_\theta, \\
X_3 = & -\sin(\theta)\,\partial_x + \cos(\theta)\,\partial_y + f\, \partial_s,\quad\quad X_4 = \partial_{f}. 
\end{split}
\end{equation}
Those horizontal vector fields endow $Q$ with a sub-Riemannian structure as explained in \cite{baspinar2020sub}. They span the \textit{horizontal tangent space} $H_{(q,z)}Q$ at each $(q,z)\in Q$. The horizontal tangent space can be thought of as the analogue of the Euclidean tangent space. In other words, differential operators such as gradient and Laplacian are defined in terms of the horizontal vector fields in the sub-Riemannian geometry. We remark that the differential operators are degenerate since the horizontal tangent space is spanned by four vector fields although it corresponds to $Q$, which is a five dimensional geometry. The horizontal integral curves are defined as the integrated curves along the horizontal vector fields given in \eqref{eq:LieAlgebra}. Albeit the degenerate character of the horizontal tangent space $H_{(q,z)}Q$, they provide the full connectivity in $Q$ due to that $X_1$ and $X_2$ do not commute as we will see below.

Nonzero commutators of the horizontal vector fields are found as
\begin{equation}
\begin{split}
[X_1,\,X_2]= &  \sin(\theta)\,\partial_x - \cos(\theta)\,\partial_y,\\
[X_2,\, X_3] = & -\cos(\theta)\,\partial_x - \sin(\theta)\,\partial_y,\\
[X_3,\,X_4] = & -\partial_s.
\end{split}
\end{equation}
The horizontal vector fields are bracket generating since
\begin{equation}\label{eq:bracketGenerating}
T_{(q,z)}Q = \operatorname{span}(X_1,\, X_2,\, X_3,\, X_4,\,[X_1,\,X_2]),
\end{equation}
for all $(q,z)\in Q$ where $T_{(q,z)}Q$ denotes the tangent space at $(q,z)\in Q$. Indeed,  \eqref{eq:bracketGenerating} shows that the horizontal vector fields fulfill H\"{o}rmander condition \cite{hormander1967hypoelliptic}. Consequently, they provide the connectivity of any two points in $Q$ through the horizontal integral curves due to the Chow-Rashevskii theorem \cite{chow1940, rashevskii1938, agrachev2019comprehensive}. This connectivity property has particular importance since it assures that any two points in the V1 sub-Riemannian model geometry $Q$ can be connected via the horizontal integral curves, which are the models of the neural connections implemented biologically in V1 and which are close approximations of the association fields at the psychophysical level.

\section{Horizontal integral curves}\label{sec:Connectivity_in_the_extended_phase_space}
The association fields were proposed to be modeled by the horizontal integral curves of $\se$ in the classical orientation sensitive framework \cite{citti2006cortical}. A similar line of thought was followed in \cite{baspinar2020sub} and it was proposed to employ the horizontal integral curves corresponding to the the five dimensional sub-Riemannian geometry $Q$ as the cortical counterparts of the association fields. The projection of those horizontal integral curves of $Q$ are the same as the projections of the horizontal integral curves of $\se$, which are shown in Figure~\ref{fig:associationFieldsIntegral}. It was conjectured in \cite{baspinar2020sub} that the horizontal integral curves of $Q$ coincide with the long range lateral connections between orientation, frequency and phase selective simple cells in V1.

\begin{figure}[htp]
\centerline{\includegraphics[scale=0.6,trim={0cm 0 0 0},clip]{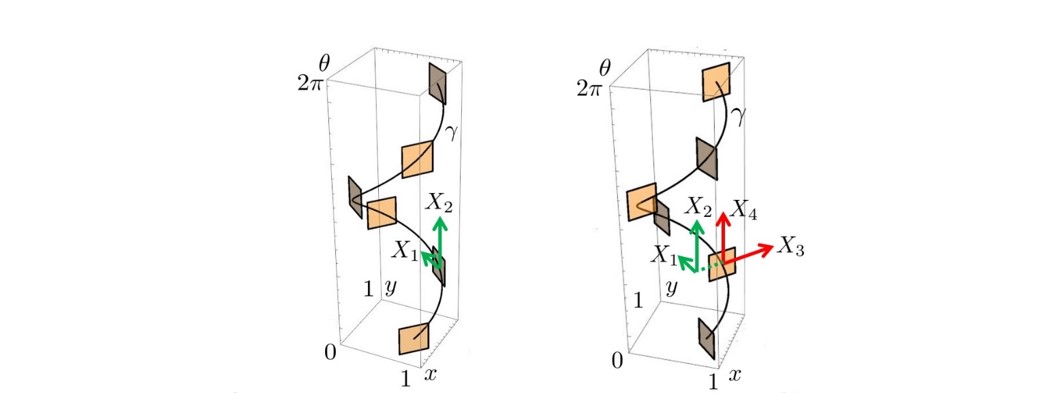}}
\caption{ A horizontal integral curve along the vector field $X_1+X_2$. It represents an orientation fiber once $f$ and $\phi$ are fixed. The tangent planes spanned by $X_1,$ $X_2$ (left) and $X_3,$ $X_4$ (right) are shown at six points on the curve.}
\label{fig:orthTan2}
\end{figure}

\begin{figure}[htp]
\centerline{\includegraphics[scale=0.325,trim={0cm 0 0 0},clip]{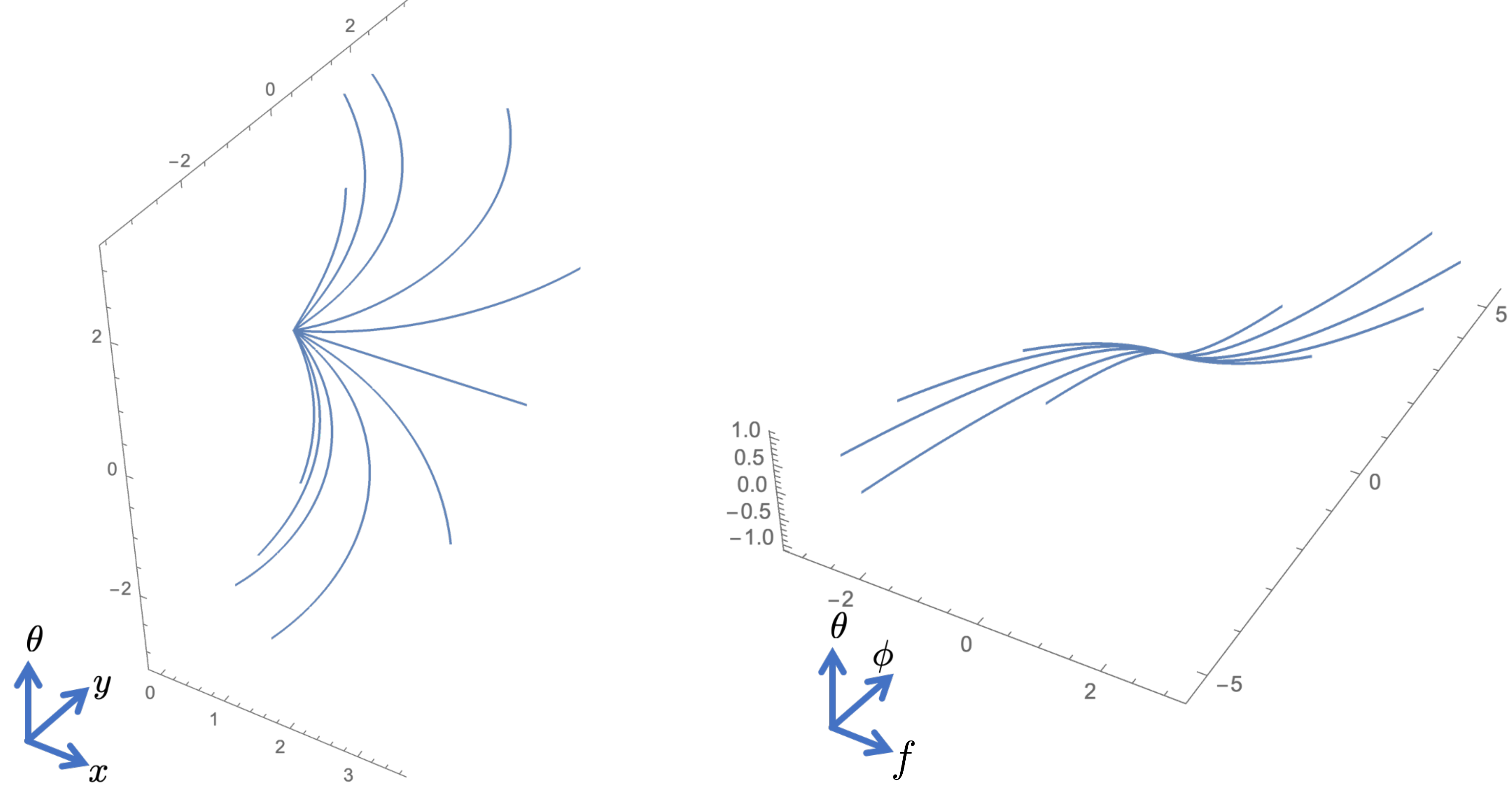}}
\caption{Horizontal integral curve fans corresponding to $X_1+c_2 X_2$ (left) and $X_3+c_4 X_4$ (right) where $c_2$ and $c_4$ are varied.}
\label{fig:integralFans}
\end{figure}

Let us denote a time interval by $\mathcal{I}=[0,T]$ with $0<T<\infty$ and consider a horizontal integral curve $(q_1,q_2,\theta,f,\phi)=\gamma:\mathcal{I}\rightarrow\mathcal{M}$ associated to the horizontal vector fields given in \eqref{eq:LieAlgebra}. We denote the initial point of $\gamma$ by $\hat{\alpha}=(\hat{q}_1,\hat{q}_2,\hat{\theta},\hat{f},\hat{\phi})$ and its velocity by $\gamma^{\prime}$. At each time instant $t\in \mathcal{I}$, the velocity is written as a vector $\gamma^{\prime}(t)\in \operatorname{span}(X_1,X_2,X_3,X_4)\big(\gamma(t) \big)$ at $\gamma(t)=(q_1(t),q_2(t),\theta(t),f(t),\phi(t))\in Q$. One way to compute the horizontal integral curves starting from the initial point $\hat{\alpha}$ is to solve the following ODE system for all $t\in \mathcal{I}$:
\begin{align}
\gamma^{\prime}(t)=
(c_1 X_1+ c_2 X_2+c_3 X_3+c_4 X_4)\vert_{\gamma(t)},
\label{eq:referredAtFanPlot}
\end{align} 
where $c_{2,3,4}$ denote the coefficients. In the activity propagation machinery which we propose here, the propagation is concentrated along a neighbourhood of the horizontal integral curves with constant coefficients $c_{2,3,4}$ in the cortical space $Q$ \cite{baspinar2020sub}; see Figures~\ref{fig:orthTan2} and~\ref{fig:integralFans}. However, the coefficients need not necessarily be constants in the generic framework of horizontal integral curves.

\FloatBarrier

\section{Sub-Riemannian diffusion in the cortical space}
\label{sec:connectivity}
The activity propagation was proposed to be modeled in terms of a sub-Riemannian diffusion process in the classical orientation sensitive $\se$ framework \cite{citti2006cortical}. This sub-Riemannian diffusion process can be interpreted as the model of the interacting neural dynamics defined in terms of the corresponding horizontal vector fields and it was applied to several biologically-inspired image processing algorithms \cite{duits2010left1, duits2010left2, boscain2012anthropomorphic, citti2016sub}. 

We follow a similar approach as in \cite{citti2006cortical} and define a sub-Riemannian diffusion procedure in the five dimensional model geometry $Q$. We denote by $\Sigma\subset Q$ the subspace in which all output responses $O_I$s are defined. These are the output responses obtained as the lifting of the two dimensional input image $I$. We will denote by $\Pi\subset \Sigma$ the subspace in which we find only the output responses corresponding to the part to be completed in the image $I$. The sub-Riemannian diffusion operator for all $(q,z)\in \Sigma$ is defined as
\[
\mathcal{L} := X_1^2 + \beta_2^2 X_2^2 +  \beta_3^2 X_3^2 + \beta_4^2 X_4^2,
\]
where $\beta_{2,3,4}$ are coefficients assuring the unit coherency in spatial, orientation, frequency and phase dimensions. The sub-Riemannian diffusion is described for all $(q,z)\in\Sigma$ by
\begin{equation}\label{eq:sRDiffusion} 
\partial_t u(q,z,t) = \mathcal{L} u(q,z,t),\quad u(q,z,0) = O_I(q,z),\quad t\in (0, T],\quad 0<T<\infty,
\end{equation}
with the corrupted region marked by the boundary conditions given by
$u(\tilde{q},\tilde{z},t) = O_I(\tilde{q},\tilde{z})$
for all $(\tilde{q},\tilde{z})\in \Sigma - \Pi$ and $t\in (0, T]$. Here $T$ denotes a sufficiently large final time and $u:Q\times [0, T] \rightarrow \mathbb{C}$ stands for the output responses evolving in time. We denote the number of orientations, frequencies and phases of the $N\times N$ image $I$ by $K$, $L$ and $M$, respectively. Then, the coefficients are found as:
\[
\beta_2 = \frac{K}{N\sqrt{2}},\quad\quad \beta_3 = \frac{L}{N\sqrt{2}},\quad\quad \beta_4 = \frac{M}{N\sqrt{2}}.
\]

We note that $\operatorname{span}(X_1, X_2)$ and $\operatorname{span}(X_3, X_4)$ define two subspaces of the horizontal tangent space $H_{(q,z)}Q$ at each point $(q,z)\in Q$. This allows us to decompose the sub-Riemannian horizontal tangent space into two components of which each one is defined by the vector fields of those two subspaces of $T_{(q,z)}Q$. Consequently, we may approximate the sub-Riemannian diffusion described by \eqref{eq:sRDiffusion} as a diffusion applied in each frequency and phase channel separately. More precisely, we apply the classical sub-Riemannian diffusion procedure defined in  $\se$ \cite{citti2006cortical, boscain2012anthropomorphic, citti2016sub} for each frequency and phase channel separately by using the $\se$ sub-Riemannian diffusion operator
\begin{equation}\label{eq:approxOperator}
\tilde{\mathcal{L}} = X_1^2 + \beta_2^2 X_2^2,
\end{equation}
where $\beta_2$ provides the unit coherency between the diffusion components in spatial and orientation dimensions. The approximate sub-Riemannian procedure is described by
\begin{equation}\label{eq:sRDiffusionApproximate}
\partial_t u(q,z,t) = \mathcal{\tilde{L}} u(q,z,t),\quad u(q,z,0) = O_I(q,z),
\end{equation}
with the boundary conditions given as $u(\tilde{q},\tilde{z},t) = O_I(\tilde{q},\tilde{z})$.
The advantage of such approximation is that we perform the diffusion procedure in $L$ three dimensional spaces instead of in a five dimensional geometry by still taking advantage of the frequency information extracted from the input image. We implement both the exact and the approximate procedures by using a simple forward Euler scheme in which the derivatives are implemented in terms of B-spline interpolated central finite differences.  

\section{Algorithm}\label{sec:algorithm}
Our algorithm is based on three steps. Once a two dimensional grayscale input image is given, the first step is lifting the image via the convolution with Gabor functions as given in \eqref{eq:liftingConv}. This provides the output responses which encode the orientation, frequency, phase values and which are represented in the five dimensional model geometry. The second step is the sub-Riemannian diffusion described by \eqref{eq:sRDiffusion}, or by~\eqref{eq:sRDiffusionApproximate} if the approximate setting is used. We integrate in time \eqref{eq:sRDiffusion} for the exact framework, and~\eqref{eq:sRDiffusionApproximate} for the approximate framework, by applying it on the output responses via iterating it with the time step $\Delta t$ until the final time $T$, at which the steady-state is reached, i.e., $\partial_t u =0$. We employ an explicit method where we use B-splined interpolated finite differences to implement the horizontal vector fields given in \eqref{eq:LieAlgebra}. The final step is to transform back the evolved output responses to the two dimensional image plane. This is achived by the inverse Gabor transform given by \eqref{eq:inverseGaborTransformExpression}. 

\subsection{Discretization of the output responses}
\label{sec:Discrete_Gabor_coeff}
We employ a uniform spatial grid to discretize the image plane such that
\begin{equation}
I[i,\,j]=I(i\Delta x,\, j\Delta y),
\end{equation}
where $i,j\in \{1,2,\dots,N\}$ with $N$ denoting the image size and $\Delta x,\Delta y\in \R^+$ denoting the pixel width. In our case, the input images are square images and $\Delta x=\Delta y=1$ in terms of pixel unit. We denote the number of samples in the orientation dimension by $K$, in the frequency dimension by $L$ and in the phase dimension by $M$. We express the distance between two adjacent samples in the orientation dimension with $\Delta\theta$, in the frequency dimension with $\Delta f$ and in the phase dimension with $\Delta s$. Discretized output response $O_I(q_{1,i},q_{2,j},\theta_k,f_l,\phi_m)$ given to $I[i,j]$ on uniform orientation, frequency and phase grids with points $\theta_k=k\, \Delta\theta$, $f_l=l\, \Delta f$ and $\phi_m=m\, \Delta s$ ($k\in \{1,2,\dots, K\}$, $l\in \{1,2,\dots,L\}$, $m\in \{1,2,\dots, M\}$) is denoted by
\begin{equation}
O_I[i,j,k,l,m]=O_I(q_{1,i},q_{2,j},\theta_k,f_l,\phi_m),
\end{equation}
where $q_{1,i}=i\Delta x$ and $q_{2,j}=j\Delta y$.

The Gabor function given by \eqref{eq:gaborExtendedFunctionExtended} is written in the discrete setting as follows:
\begin{equation}
\Psi_{[i,j,k,l,m]}[\tilde{i},\tilde{j},\tilde{n}]=\Psi_{(q_{1,i},\,q_{2,j},\,\theta_k,\,f_l,\,\phi_m)}(x_{\tilde{i}},y_{\tilde{j}},s_{\tilde{n}}),
\end{equation}
where $\tilde{i},\tilde{j}\in\{1,2,\dots, N \}$, $\tilde{k}\in\{1,2,\dots, K \}$, $\tilde{n}\in \{1,2,\dots, M \}$,
for each orientation $\theta_k$, frequency $f_l$ and phase $\phi_m$. We fix $s_{\tilde{n}}=0$ and express the discretized cell response obtained from the input image $I[i,j]$ via the lifting described by the discrete Gabor transform as follows:
\begin{equation}\label{eq:discreteLift}
O_I[i,j,k,l,m]=\SUM_{\tilde{i},\tilde{j}}\Psi_{[i,j,k,l,m]}[\tilde{i},\tilde{j},0]\,I[\tilde{i},\tilde{j}].
\end{equation}
We discretize the time interval by $V\in\mathbb{N}^+$ samples, and denote it by $h_v$. Here $h_v$ is the time instant $h_v=v\,\Delta t$ with $\Delta t$ satisfying $T=V\,\Delta t$ and $v\in\{1,2,\dots,V\}$. Discretized evolving output response is written as
\begin{equation}
U[i, j, k, l, m, v] = u(q_{1,i}, q_{2,j}, \theta_k, f_l, \phi_m, h_v)
\end{equation}
Finally, the discrete inverse transform applied via a normalized kernel $\bar{\Psi}$ on the evolved output responses until the final time $T$ gives the completed two dimensional image $I_T$, which is found as follows:
\begin{align}\label{eq:discreteInverseTransfrom}
I_T[i,j] = \sum_{\tilde{i},\tilde{j},\tilde{k},\tilde{m}} \sqrt{f_{\tilde{l}}}\,\sum_{\tilde{l}}  U_T[\tilde{i},\tilde{j},\tilde{k},\tilde{l},\tilde{m}]\bar{\Psi}_{[\tilde{i},\tilde{j},\tilde{k},\tilde{l},\tilde{m}]}[i,j,0].
\end{align}

\subsection{Explicit scheme with finite differences}
Here we provide the explicit numerical scheme which we employ to iterate the exact and approximate frameworks given in \eqref{eq:sRDiffusion} and \eqref{eq:sRDiffusionApproximate}, respectively. Our motivation for choosing an explicit scheme rather than an implicit scheme is that the latter requires large memory and computational power in our multidimensional framework.

We follow \cite{unser1999splines, duits2010left2} to implement the horizontal vector fields given in \eqref{eq:LieAlgebra} via B-spline interpolated central finite differences. The interpolation takes place on a uniform grid. It is needed since the horizontal vectors are not always aligned with the spatial grid point samples. B-spline interpolation is based on the coefficients $b(i,j)$
\begin{equation}
\operatorname{sp}(x,y)=\SUM_{i,j\in Z}b(i,j)\rho(x-i,y-j).
\end{equation}
The coefficients are determined such that the spline polynomial $\operatorname{sp}(x,y)$, together with the B-spline basis functions $\rho(x-i, y-j)$, coincides with the horizontal derivatives at the grid points. For example, the condition $\operatorname{sp}(i\Delta x, j\Delta y)=X_1O^{I}[i,j,k,l,m]$ must be satisfied once the correct coefficients $b$ are determined; see \cite{unser1999splines, franken2008enhancement, duits2010left2} for more details.

\begin{figure}[htp]
\centerline{\includegraphics[scale=0.3,trim={0cm 0cm 0cm 0cm},clip]{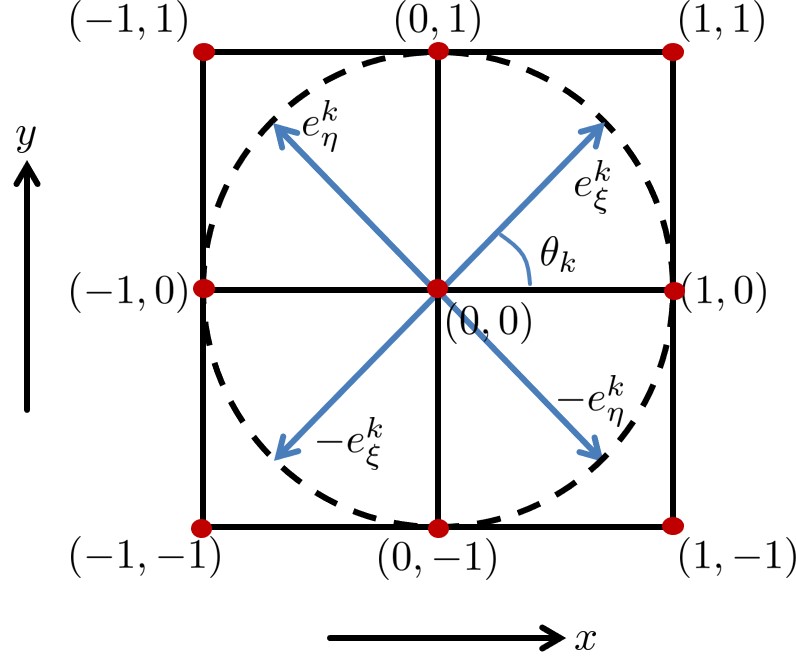}}
\caption{Illustration of the vectors $e^k_{\xi}$ and $e_{\eta}^k$ at $(0,0)$ with $\Delta x=\Delta_y=1$. The figure was modified and adapted from \cite{duits2010left2}.}
\label{fig:finiteDifferenceScheme2}
\end{figure}

We define
\begin{align}
\begin{split}
e^k_{\xi}:= & (\Delta x\cos(\theta_k),\,\Delta y\sin(\theta_k)),\\
e^k_{\eta}:= & (-\Delta x\sin(\theta_k),\,\Delta y\cos(\theta_k)),
\end{split}
\end{align}
whose illustrations corresponding to $\Delta x = \Delta y =1$ case are given in Figure~\ref{fig:finiteDifferenceScheme2}. We abuse the notation to denote the evolving output responses:
\begin{equation}
U = U[i,j,k,l,m, v] = u(q_{1,i}, q_{2,j}, \theta_k, f_l, \phi_m, h_v),
\end{equation}
and then we write the central finite differences of the second order horizontal derivatives as

\begin{align}
\label{eq:fdScheme2}
\begin{split}
X_1 X_1 U\approx \frac{1}{(\Delta x)^2}  &  \Big( u(q+e^k_{\xi},\theta_k,f_l,\phi_m)-2 u(q,\theta_k,f_l,\phi_m)\\ & + u(q-e^k_{\xi},\theta_k,f_l,\phi_m)\Big),\\
X_2X_2 U\approx \frac{1}{(\Delta \theta)^2}  &  \Big(u(q,\theta_{k+1},f_l,\phi_m)-2 u(q,\theta_{k},f_l,\phi_m)\\   & + u(q,\theta_{k-1},f_l,\phi_m)\Big),\\
X_3X_3 U \approx \frac{1}{(\Delta x)^2}  & \Big( u(q+e^k_{\eta},\theta_k,f_l,\phi_m)-2 u(q,\theta_k,f_l,\phi_m)+ u(q-e^k_{\eta},\theta_k,f_l,\phi_m) \Big)\\
&+\frac{f_l^2}{(\Delta s)^2}\Big( u(q,\theta_{k},f_l,\phi_{m+1})-2 u(q,\theta_{k},f_l,\phi_m)\\
& + u(q,\theta_{k},f_l,\phi_{m-1})\Big) +\frac{f\cos(\theta_k)}{2\Delta s \Delta_x}\Big (u(q+e^k_{\eta}, \theta_{k},f_l,\phi_{m+1})\\
& -u(q-e^k_{\eta},\theta_{k},f_l,\phi_{m+1})
-u(q+e^k_{\eta},\theta_{k},f_l,\phi_{m-1})\\
& +u(q-e^k_{\eta},\theta_{k},f_l,\phi_{m-1})\Big)-\frac{f \sin(\theta_k)}{2\Delta s \Delta_x}\Big(u(q+e^k_{\xi}, \theta_{k},f_l,\phi_{m+1})\\
& -u(q-e^k_{\xi},\theta_{k},f_l,\phi_{m+1})
-u(q+e^k_{\xi},\theta_{k},f_l,\phi_{m-1})\\
& + u(q-e^k_{\xi},\theta_{k},f_l,\phi_{m-1})\Big) +\frac{f^2}{2\Delta s} \Big(u(q,\theta_{k},f_l,\phi_{m+1})\\ 
& - 2 u(q,\theta_{k},f_l,\phi) +u(q,\theta_{k},f_l,\phi_{m-1})\Big ),\\
X_4 X_4 u[i,j,k,l,m]\approx \frac{1}{(\Delta f)^2} & \Big( u(q,\theta_{k},f_{l+1},\phi_m)-2 u(q,\theta_{k},f_l,\phi_m)+u(q,\theta_{k},f_{l-1},\phi_m) \Big).
\end{split}
\end{align}
Finally we write the discretized numerical iteration for \eqref{eq:sRDiffusion} and \eqref{eq:sRDiffusionApproximate} as follows:
\begin{equation}\label{eq:discreteGeneralExpression}
\begin{split}
U[i,j,k,l,m, v]= & u(q_{i,1},q_{j,2},\theta_{k},f_l,\phi_m, h_v) \\
= & u{(q_{i,1},q_{j,2},\theta_{k},f_l,\phi_m, h_{v-1})}+\Delta t \,\bar{\mathcal{L}} u(q_{i,1},q_{j,2},\theta_{k},f_l,\phi_m, h_{v-1}),
\end{split}
\end{equation}
where $\bar{\mathcal{L}}$ represents the discretized version of either $\mathcal{L}$ or $\tilde{\mathcal{L}}$, depending on which one between exact and approximate frameworks is chosen. The discretization is achieved by replacing the second order horizontal derivatives with their discrete versions given in~\eqref{eq:fdScheme2}.

\subsection{Pseudocode of the algorithm}
We denote the processed two dimensional image at the final time $T$ by $I_T$ and the evolving discrete output responses at the time instant $v\,\Delta t $ by $U_v$. Then we provide a general scheme of the exact completion algorithm as follows:
\begin{algorithm}
 \caption{Completion algorithm
pseudocode.}
 \label{algo:pseudocode}
\SetAlgoLined
\KwData{$N\times N$ input grayscale image $I$ \\ \textbf{Parameters}: $\sigma, \beta_2, \beta_3, \beta_4, T, K, L, M, \Delta t,, \Delta\theta, \Delta f, \Delta s$, \texttt{tol}}
\KwResult{Processed image $I_T$}
 Compute the lifting $O_I$ via \eqref{eq:discreteLift}\;
 Initialize the iteration index $v\leftarrow 0$\;
 \Repeat{$({\|U_v - U_{v-1}\|_{\mathbb{L}^2}}/{\|U_v\|_{\mathbb{L}^2}}) < \texttt{tol}$}{
  $v\leftarrow v+1$\;
  Compute the right hand side of~\eqref{eq:discreteGeneralExpression}\;
  Update $U_v$ via~\eqref{eq:discreteGeneralExpression}\;
 }
 Inverse transform the evolved output responses $U_T$ to obtain the completed image $I_T$ via~\eqref{eq:discreteInverseTransfrom}.
\end{algorithm}
\newline The detailed Matlab\,\textsuperscript{\textregistered} code can be found in the following link: \url{https://drive.google.com/file/d/1YET47AxYo5_FQs3ObREAqvQp2KjuKh7n/view} ({accessible starting from 29 October 2021}).

\section{Numerical Experiments}
\begin{figure}
\centering
 \includegraphics[scale=0.75]{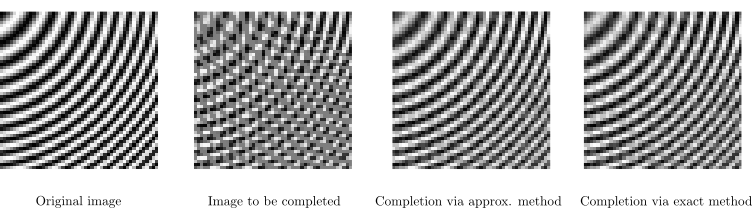}
\caption{Completion of arcs with sinusoidal pattern. Left: original image. Middle left: corrupted image with occluding arcs. Middle right: completed image via the approximate method. Right: completion via the exact method.}
\label{fig:completionSin}
\end{figure}

\begin{figure}
\centering
 \includegraphics[scale=0.733]{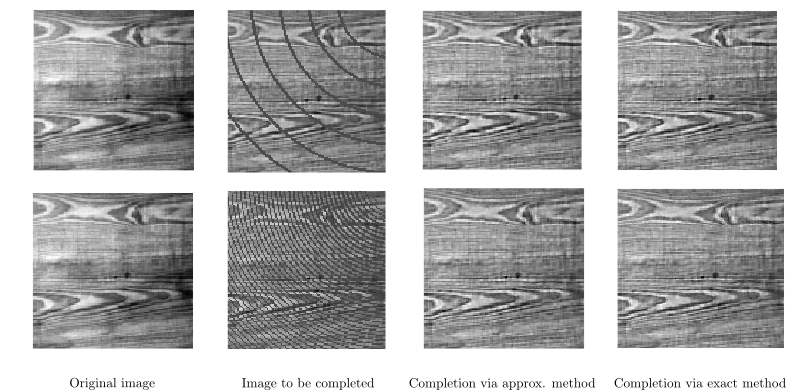}
\caption{Completion of an occluded real texture image by two different arc patterns on the top and bottom rows. Left: Original image taken from. Middle left: Image with occluding arcs. Middle right: Completed image via the approximate framework. Right: Completed image via the exact framework.}
\label{fig:completionAllWood}
\end{figure}

\begin{figure}
\centering
 \includegraphics[scale=0.723]{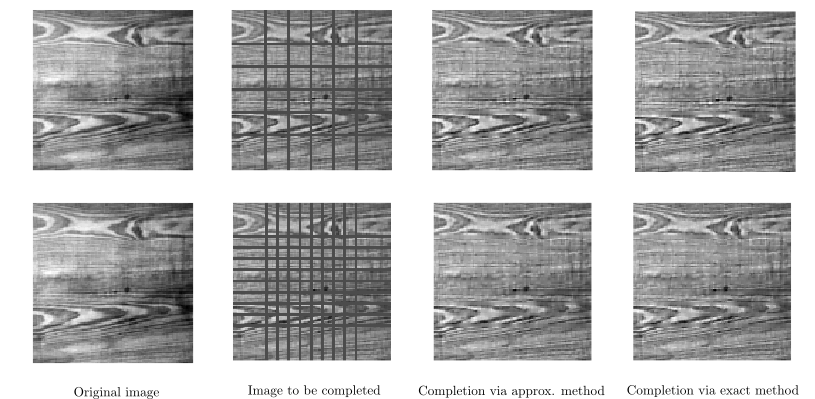}
\caption{Completion of a real texture image occluded by two different line patterns on the top and bottom rows. Left: original image. Middle left: image with occluding vertical and horizontal lines. Middle right: completed image via the approximate framework. Right: completed image via the exact framework.\bigskip}
\label{fig:completionAllWood2}
\end{figure}

\begin{figure}
\centering
 \includegraphics[scale=1]{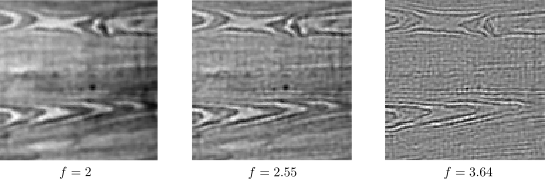}
\caption{Single frequency completion associated to the bottom row of Figure~\ref{fig:completionAllWood}}
\label{fig:singleFrequencyCompletion}
\end{figure}

\begin{figure}
\centering
 \includegraphics[scale=1]{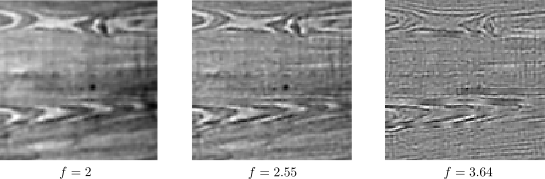}
\caption{Single frequency completion associated to the bottom row of Figure~\ref{fig:completionAllWood2}}
\label{fig:singleFrequencyLineCompletion}
\end{figure}

\begin{figure}
\centering
 \includegraphics[scale=0.75]{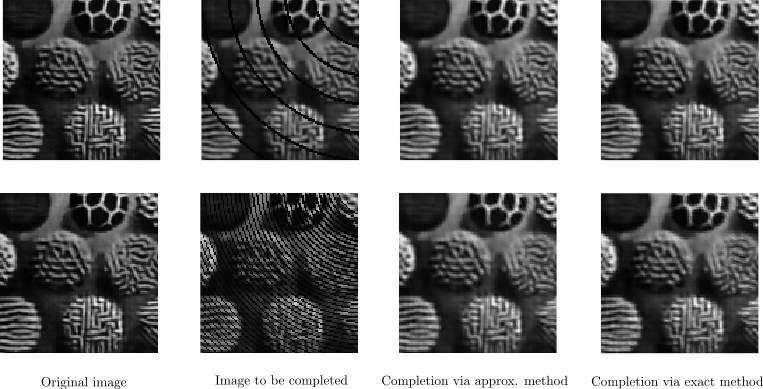}
\caption{Completion of an occluded real texture image by two different arc patterns on the top and bottom rows. Left: original image taken from \cite{kimmel2000images}. Middle left: image with occluding arcs. Middle right: completed image via the approximate framework. Right: completed image via the exact framework.}
\label{fig:completionArcsTexture}
\end{figure}

\begin{figure}
\centering
 \includegraphics[scale=0.75]{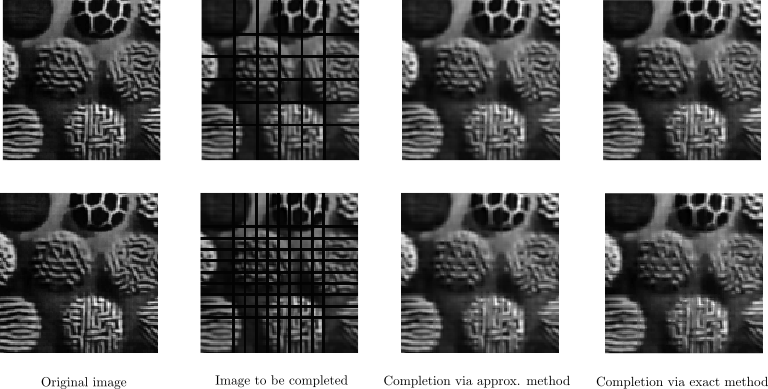}
\caption{Completion of an occluded real texture image by two different line patterns on the top and bottom rows. Left: original image taken from \cite{kimmel2000images}. Middle left: image with occluding arcs. Middle right: completed image via the approximate framework. Right: completed image via the exact framework.}
\label{fig:completionLinesTexture}
\end{figure}

\begin{figure}
\centering
 \includegraphics[scale=1]{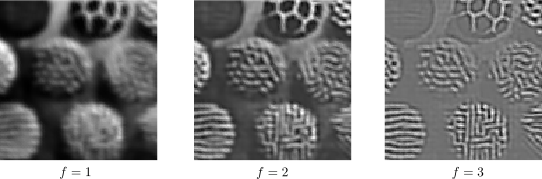}
\caption{Single frequency completion associated to the bottom row of Figure~\ref{fig:completionArcsTexture}}
\label{fig:completionARcsTextureSingle}
\end{figure}

\begin{figure}
\centering
 \includegraphics[scale=1]{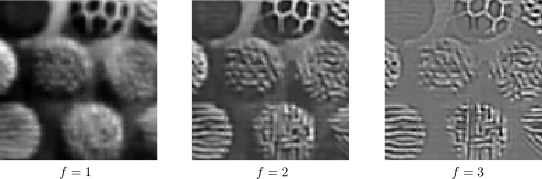}
\caption{Single frequency completion associated to the bottom row of Figure~\ref{fig:completionLinesTexture}}
\label{fig:singleFrequencyTextureLineCompletion}
\end{figure}

\begin{figure}
\centering
 \includegraphics[scale=0.75]{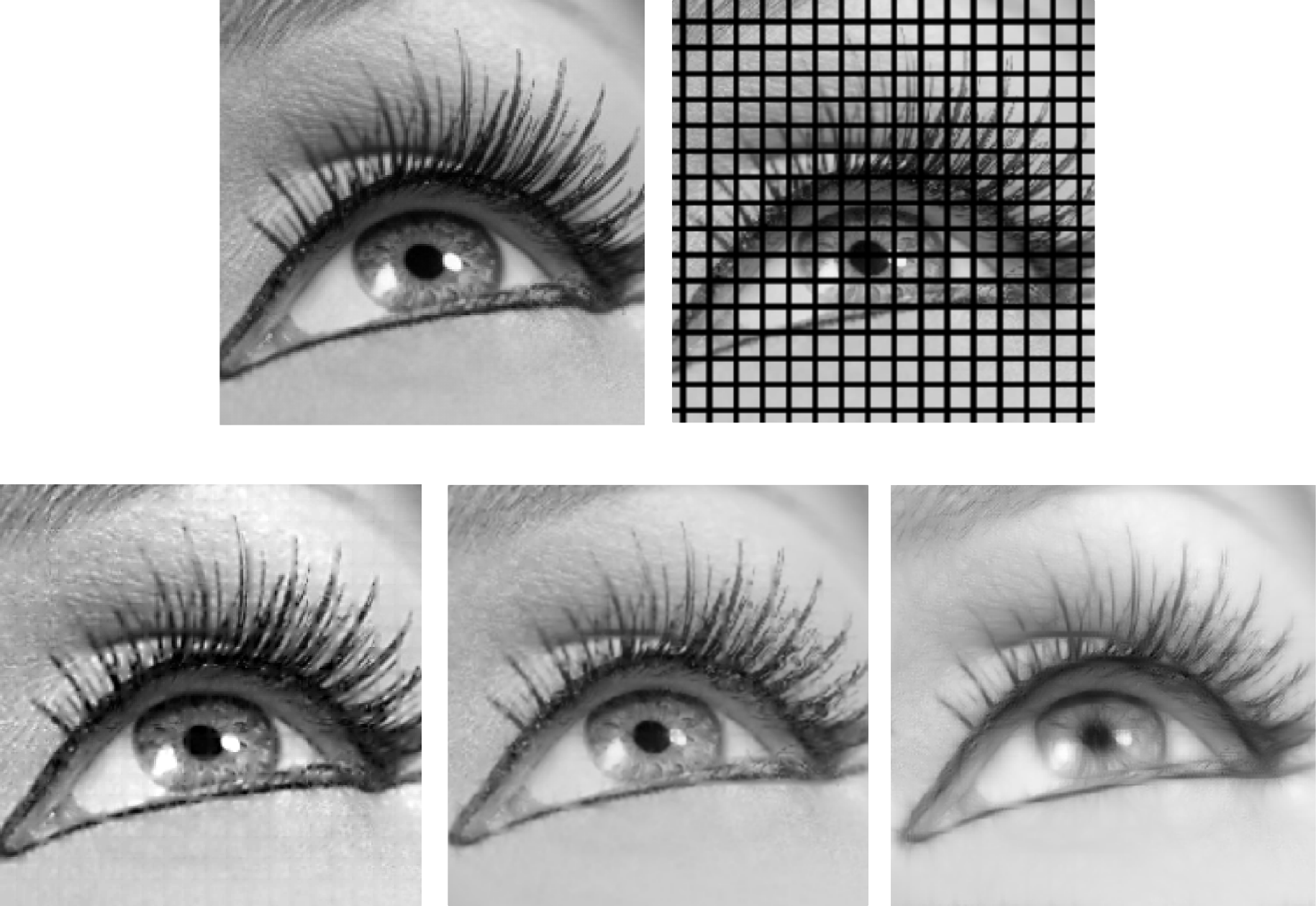}
\caption{Top left: test image from \cite{boscain2018highly}. Top right: image to be completed. Bottom left: completion via our method. Bottom middle: completion via the method in \cite{citti2016sub}. Bottom right: completion via the method in \cite{boscain2014hypoelliptic}. }
\label{fig:comparisonOthers}
\end{figure}

We choose $\sigma=2$ as the scale of the localizing Gaussian given in \eqref{eq:gaborExtendedFunctionExtended} for all experiments. We use $128\times 128$ grayscale images in the experiments related to Figures~\ref{fig:completionSin}-\ref{fig:singleFrequencyTextureLineCompletion} and $256\times 256$ grayscale image in the experiment corresponding to Figure~\ref{fig:comparisonOthers}. We use $\theta \in \{0, \frac{\pi}{32}, \frac{2\pi}{32}, \frac{3\pi}{32},\dots, \frac{31\pi}{32} \}$, $\phi \in \{0, \frac{\pi}{8}, \frac{2\pi}{8}, \frac{3\pi}{8}, \frac{4\pi}{8} \}$ and $\Delta t = 0.1$ in all the experiments. 

Our first results are obtained by using an artificial test image as shown in Figure~\ref{fig:completionSin}. In the test image, we have arcs of circles centered at the top left corner and with different radii, where a sinusoidal function whose frequency increases linearly as the radius increases. The arcs with sinusoidal pattern are occluded by zero valued arcs belonging to the circles which are centered at the top right corner. We apply our completion procedure with $T=10$ and $f\in \{2, 2.5455, 3.0909,\dots, 8   \}$. We observe that both the approximate and exact frameworks show a similar performance and they provide proper completion.

In Figure~\ref{fig:completionAllWood}, we present two cases of our completion algorithm applied to a real texture image which is occluded by arcs corresponding to circles with different radii. We perform our completion procedure with $f\in \{2.00, 2.55,    3.09, 3.64, 4.18, 4.73, 5.27, 5.82, 6.36, 6.91, 7.45, 8.00\}$. Here $T=10$ for the upper row and $T=5$ for the bottom row. We use the same parameters but now $T=15$ in Figure~\ref{fig:completionAllWood2} on the same texture image but now occluded by vertical and horizontal bars. In such set of test images, the challenge is that the bars cross each other. Our completion algorithms provide proper completion of such crossing areas. We observe that both the approximate and the exact algorithms are able to complete in all cases the occluded parts in Figures~\ref{fig:completionAllWood} and~\ref{fig:completionAllWood2}. 

In order to see the impact of using multiple frequencies in the completion process via the sub-Riemannian diffusion, we performed the sub-Riemannian diffusion by using only one single frequency; see Figures~\ref{fig:singleFrequencyCompletion} and~\ref{fig:singleFrequencyLineCompletion}. In other words, $f$ is fixed to a constant both in the lifting procedure and in the sub-Riemannian diffusion. In Figure~\ref{fig:singleFrequencyCompletion}, the same corrupted image given in the bottom row of Figure~\ref{fig:completionAllWood} was used. The paramaters are the same as in the bottom row results of Figure~\ref{fig:completionAllWood}. We used $f=2,\; 2.55,\; 3.64$ from left to right. We observe that the performance of single frequency framework is visibly lower than the multi-frequency framework whose results are presented in Figure~\ref{fig:completionAllWood} due to the presence of a wide range of spatial frequencies in the image to be completed. We observe the same also in the single frequency results presented in Figure~\ref{fig:singleFrequencyLineCompletion}, which are obtained by using the bottom row corrupted image given in Figure~\ref{fig:completionAllWood2}. Finally, we note that the inverse Gabor transform in those single frequency experiments is not well defined since the information corresponding to the other frequency components in the input image is lost in the lifting procedure. In other words, Parseval formula associated to multi-frequency Gabor transform does not hold any longer \cite[Equation 2]{duits2013evolution}. Therefore, we simply project the processed output responses onto the image plane via a sum over orientation and phase components in the single frequency results shown in Figures~\ref{fig:singleFrequencyCompletion} and~\ref{fig:singleFrequencyLineCompletion}.

In Figures~\ref{fig:completionArcsTexture} and~\ref{fig:completionLinesTexture}, we perform the same type of experiments as in Figures~\ref{fig:completionAllWood} and~\ref{fig:completionAllWood2}, respectively, but now with $f\in \{1.00, 1.27, 1.55, 1.82, 2.09, 2.36, 2.64, 2.91, 3.18, 3.45, 3.73, 4.00\}$ and by employing a different type of texture image. Here $T=10$ for both rows in Figure~\ref{fig:completionArcsTexture}, and $T=15$ for both rows in Figure~\ref{fig:completionLinesTexture}. In Figure~\ref{fig:completionARcsTextureSingle}, we performed the single frequency experiments by using the occluded image found in the bottom row of Figure~\ref{fig:completionArcsTexture}. Similarly, in Figure~\ref{fig:singleFrequencyTextureLineCompletion}, we performed the same single frequency experiments but now using the occluded image given in the bottom row of Figure~\ref{fig:completionLinesTexture}. Similarly to the case of the previous real test image, single frequency completion cannot perform a proper completion due to the loss of the information corresponding to the other frequencies present in the input image.

Finally, in Figure~\ref{fig:comparisonOthers}, we compare the result of our algorithm to the results obtained by applying the algorithms explained in \cite{citti2016sub} and \cite{boscainHypoelliptic2014}. The method proposed in \cite{citti2016sub} is defined in the classical model framework $\se$ and it combines the $\se$ sub-Riemannian diffusion with a concentration mechanism resulting in a diffusion driven motion by curvature in $\se$. The algorithm explained in \cite{boscainHypoelliptic2014} uses a semidiscrete version of the classical model geometry $\se$, and this allows to perform completion via the integration of parallelisable finite set of Mathieu-type diffusions combined with a dynamical restoring mechanism. The main difference between our method and those previously proposed algorithms is that our model uses a higher order sub-Riemannian geometry, this allows to take into account frequency and phase information as well. Moreover, we do not combine in our framework the diffusion procedure with a concentration or dynamical restoring mechanism. We see in Figure~\ref{fig:comparisonOthers} that our algorithm produces comparable completion results to the other two methods. We observe that our algorithm is able to preserve the contextual information better than the other two, especially the high frequency structures, thanks to the use of multiple frequencies and the exact inverse Gabor transform. The trade off is that in the corresponding stationary state, especially in the low frequency parts such as the below eye region, the completion is weaker compared to the other two methods. This supports that our algorithm is better adapted to the texture images, such as the ones in Figures~\ref{fig:completionAllWood}, \ref{fig:completionAllWood2}, than to the natural images such as the one given in Figure~\ref{fig:comparisonOthers}. In our simulation in Figure~\ref{fig:comparisonOthers}, we used $f\in\{1.50, 2.09, 2.68, 3.27, 3.86, 4.45, 5.05, 5.64, 6.23, 6.82, 7.41, 8.00 \}$ and $T = 50$.    

\FloatBarrier

\section{Conclusion}
In this work, we presented a completion algorithm which uses multiple frequency and phase channels to take advantage of the spatial frequency and phase information of a given two dimensional grayscale image. The algorithm consists of three mechanisms: feature extraction, sub-Riemannian diffusion and inverse transform. The first one is a linear filtering of the given input image with the Gabor filter banks. The filtering encodes the visual feature values in the output responses, i.e., orientation, frequency and phase values, of each pixel in the two dimensional input image. The oputput responses are represented in the five dimensional sub-Riemannian model geometry. The second mechanism, which is the sub-Riemannian diffusion, models the activity propagation between the simple cells in V1. It is concentrated in a neighbourhood along the horizontal integral curves corresponding to the sub-Riemannian model geometry. Those horizontal integral curves are conjectured to be good approximations of the neural connections in V1 \cite{baspinar2020sub}. Once they are projected onto the two dimensional image plane, their projections overlap closely with the association fields as was shown in Figure~\ref{fig:associationFieldsIntegral}. Resulting from the sub-Riemannian diffusion, subjective contours of amodal completion are reconstructed in the five dimensional model goemetry. Finally, the inverse Gabor transform provides the representation of the evolved output responses, therefore the reconstructed subjective contours together with the rest of the lifted image, on the two dimensional image plane. This final result is the completed image in which the initially occluded parts are revealed.  

One of the novelty of the algorithm is that it is not only an image completion algorithm but it takes into account neurophysiological and psychophysical orientation, frequency and phase constraints observed in the visual cortex. Therefore, it should not be considered as an highly specialized image processing algorithm such as those found in medical imaging, radar imaging, robotics and computer vision. It should be considered rather as an algorithm compatible with a natural geometry which was \textit{derived} from one of the first step mechanisms of the mammalian visual perception, from receptive profile. In other words, it reflects the cortical architecture. Moreover, it uses multi-frequency and phase channels, which was not the case in the previously proposed completion algorithms using the classical orientation sensitive $\se$ sub-Riemannian framework \cite{citti2006cortical, citti2016sub} and its variant \cite{boscain2014hypoelliptic}. This allows our algorithm to employ the inverse Gabor transform instead of projecting the processed output responses onto the two dimensional image plane to provide the completed final image; this was not possible in the aforementioned previous methods \cite{citti2006cortical, boscain2014hypoelliptic, citti2016sub}. This provides good preservation of the contextual information (contours, edges etc.) with different frequencies in the input image. 

One of the interesting aspects for future work is to consider a concentration mechanism. It is possible to embed in the proposed completion algorithm a similar concentration mechanism to the one presented in \cite{citti2006cortical}, but now with a concentration in each frequency and phase channel. Moreover, the proposed completion algorithm uses the same model geometry as the enhancement algorithm which was presented in \cite{baspinar2020sub}. Another interesting future work is to combine those two algorithms to perform both completion and enhancement at the same time by employing orientation, frequency and phase information existing in a two dimensional input image. Finally, the study of an analytical solution in a similary way as was done in $\se$ \cite{duits2008explicit, zhang2016numerical}, but this time for the sub-Riemannian diffusion defined in the five dimensional model geometry, could provide new techniques to perform the completion task, as well as many other image processing applications in the model geometry. This would open new interesting questions especially at theoretical level. 

\section*{Acknowledgements}
I would like to thank to Giovanna Citti and Alessandro Sarti for the inspiring discussions. I hereby acknowledge that I was supported by the Human Brain Project (European Union grant H2020-945539).

\bibliographystyle{ieeetr}
\bibliography{meetingNotes_Bib}

\end{document}

%% file: ABD_shared.tex
\usepackage{lipsum}
\usepackage{amsfonts}
\usepackage{graphicx}
\usepackage{epstopdf}
\usepackage{algorithmic}
\ifpdf
  \DeclareGraphicsExtensions{.eps,.pdf,.png,.jpg}
\else
  \DeclareGraphicsExtensions{.eps}
\fi

\usepackage{enumitem}
\setlist[enumerate]{leftmargin=.5in}
\setlist[itemize]{leftmargin=.5in}
\usepackage{todonotes} 
\usepackage{amsmath,amssymb}
\usepackage{placeins}
\usepackage{cite}
\usepackage{float}

\newsiamremark{remark}{Remark}
\newsiamremark{hypothesis}{Hypothesis}
\crefname{hypothesis}{Hypothesis}{Hypotheses}
\newsiamthm{claim}{Claim}

\newcommand{\R}{\mathbb{R}}
\newcommand{\SUM}{\sum\limits}

\newcommand{\expp}{\operatorname{e}}

\headers{Multi-frequency image completion via a biologically-inspired model }{E. Baspinar}

\title{Multi-frequency image completion via a biologically-inspired sub-Riemannian model with frequency and phase}

\author{Emre Baspinar\thanks{CNRS/NeuroPSI, Paris-Saclay Campus CEA, France  (\email{emre.baspinar@cnrs.fr}).}}

\usepackage{amsopn}
